\documentclass[10pt,twocolumn,letterpaper]{article}

\usepackage{cvpr}
\usepackage{times}
\usepackage{epsfig}
\usepackage{graphicx}
\usepackage{tabularx}
\usepackage{colortbl}
\usepackage{amsmath}
\usepackage{algorithm}
\usepackage{appendix}
\usepackage[noend]{algpseudocode}
\usepackage{hhline}
\usepackage{amssymb}
\usepackage[usenames,dvipsnames,svgnames,table]{xcolor}
\newcommand\floor[1]{\lfloor#1\rfloor}
\definecolor{cmcolor}{rgb}{0,0.6,0}
\newcommand{\comment}[1]{}
\newcommand{\mat}[1]{\mathbf{#1}}
\newcommand{\red}[1]{\textcolor{black}{{{} #1}}}
\renewcommand{\algorithmicrequire}{\textbf{Input:}}
\renewcommand{\algorithmicensure}{\textbf{Output:}}

\usepackage[pagebackref=true,breaklinks=true,letterpaper=true,colorlinks,bookmarks=false]{hyperref}

\cvprfinalcopy 

\def\cvprPaperID{513} 

\ifcvprfinal\fi
\begin{document}

\title{Multi-velocity neural networks for gesture recognition in videos}

\author{Otkrist Gupta\\
{\tt\small otkrist@mit.edu}
\and
Dan Raviv\\
{\tt\small raviv@mit.edu}\\
Massachusetts Institute of Technology\\
Cambridge, MA\\
\and
Ramesh Raskar\\
{\tt\small raskar@media.mit.edu}
}
\teaser{}

\maketitle
\begin{abstract}

\red{We present a new action recognition deep neural network which adaptively learns the best action velocities in addition to the classification. While deep neural networks have reached maturity for image understanding tasks, we are still exploring network topologies and features to handle the richer environment of video clips. Here, we tackle the problem of multiple velocities in action recognition, and provide state-of-the-art results for gesture recognition, on known and new collected datasets. We further provide the training steps for our semi-supervised network, suited to learn from huge unlabeled datasets with only a fraction of labeled examples.}
\end{abstract}

\section{Introduction}
Nonverbal communication is a key factor in the interaction between individuals, replacing or amplifying spoken words. Our body language, voice pitch, intonation and volume, movement of our pupils or our chronemics choices are just a few examples emphasizing the richness of human communication skills \cite{mehrabian1974approach}. A special subset of nonverbal interactions, explored in this paper, is based on facial expressions. Their perception initiates rapid cognitive processes in the brain and have both communicative and reflexive components \cite{frith2009role}. They  assist in verbal communication by providing context to what we are saying, making their recognition important for studying social interactions.

\begin{figure*}
\begin{center}
\includegraphics[width=0.95\linewidth]{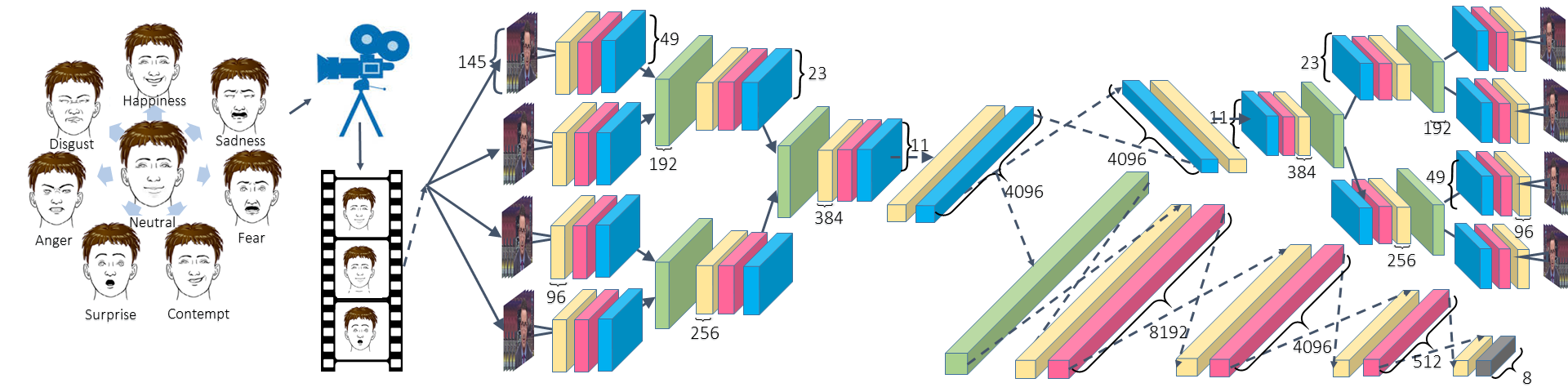}
\end{center}
   \caption{We learn 7 different facial emotions from short (about 1 sec, 25 frames) video clips. Our prediction system is based on slow temporal fusion neural network, trained by hybridization of autoencoding a huge collected dataset and a loss prediction on a small set of labeled gestures.}
\label{fig:teaser}
\end{figure*}

We use computers on a daily basis, interacting with artificial agents in increasing number of tasks. Advances in speech and natural language processing have presented us with personalized smart agents  \cite{siewiorek2008application,freed2008radar}, while examining our facial gestures has provided a richer set of tools for improving human computer interaction \cite{cowie2001emotion}. In the last two decades we have seen several attempts to automate the way computers respond towards human emotions \cite{klein1999computer,cerezo2007interactive,andre2000exploiting}, where the ultimate goal is to create humanoid robots which can blend in the environment \cite{brown2014meet}.

Unfortunately this task is hard to solve, and current state-of-the-art results are far from satisfactory. Recent advances in machine learning have shown that if we provide a neural network with enough samples, it can learn very complex structures \cite{hinton2006fast}. Today hard tasks in computer vision, such as labeling images, recognizing objects and faces or classifying videos  have become a feasible task for computers which can now provide competitive results to humans and sometimes even outperform them \cite{ranzato2011deep,taigman2014deepface}.


In this paper we focus on facial gesture recognition from videos using deep neural networks. We tackle the problem of a small size labeled dataset, and present a new layer which compensates for velocity changes in the time domain. We compare our methods to multiple techniques and datasets, as well as presenting our own collected data, and we report state-of-the-art results in almost every category. \red{We hope to empower researchers in this area by providing them with a huge dataset which can help them build even bigger and better deep neural networks, while eliminating the need to spend several months required to acquire a dataset of such proportions, and for some living under limited resources acquiring such a dataset can be prohibitively impossible.}
We summarize our contributions by:
\subsection{Contributions}

\begin{enumerate}

\item \red{To the best of out knowledge, we have built \textbf{the largest face video dataset} to date comprising of \textbf{162 million} face images with facial landmark labels (\textbf{7.8 billion annotations}) contained in 6.5 million video clips, and 2777 videos which have been labeled for seven emotions. The dataset will be made public for research purposes.}

\item \red{We develop a multi-velocity autoencoder architecture using \textbf{new multi-velocity} layers for generating velocity-free deep motion features.}

\item We report state of the art results for video gesture recognition using spatio-temporal convolutional neural networks.

\item We introduce a new topology and protocol for semi-supervised learning, where the number of labeled data points is only a fraction of the entire dataset.

\end{enumerate}

\section{Related Work}

Machine learning techniques such as Support Vector Machines have been used for facial expression recognition given the movement of facial fiducial points \cite{kotsia2007facial,michel2003real,shan2005robust,dhall2011emotion} achieving real time performance  \cite{ren2014face}.  Many of these techniques involve a pipeline with multiple phases - face detection and alignment, feature extraction/landmark localization and classification as the final step. Other interesting approaches \cite{chen20153d,walecki2015variable,presti2015using,vieriu2015facial} we should mention are based on temporal features \cite{liu2014learning,wang2013capturing}, and multiple kernels \cite{liu2014combining}, action units \cite{zhao2015joint,senechal2015facial}, as well as emotion recognition from speech \cite{nwe2003speech,schuller2004speech}.
We will compare our method against some of those approaches in section \ref{experimentsandresults}.

Recently, deep neural nets have been shown to perform well on classification tasks on images and videos, outperforming most traditional learning systems. One of the most interesting results was presented three years back on a large scale dataset (LSVRC 2011), where a deep convolutional net outperformed all other methods by far  \cite{krizhevsky2012imagenet}. With advances in convolutional neural nets, we have seen neural nets applied to video classification \cite{karpathy2014large,tran2014learning} and even facial expression recognition \cite{abidin2012neural,gargesha2002facial} but these networks were not deep enough or used other feature extraction techniques like PCA or Fisherface.

Training a neural net normally requires a large labeled dataset which is hard to obtain using reasonable resources. Providing high quality results when only a small part of the data is labeled is an interesting problem referred to as semi-supervised learning. In \cite{lee2013pseudo} the authors pre-trained the system using pseudo labels, while in \cite{weston2012deep,kingma2014semi} they embedded the data in a low dimensional space. Very recently superior results have been shown \cite{liu2014facial,kahou2013combining,jung2015joint,he2015multimodal,kahou2015emonets} using deep neural nets to combine labels and un-labeled data in the same package. In this paper we follow those guidelines and train from start-to-end a hybrid system composed of autoencoders for unlabeled data and additional loss function for the classification tasks.

\section{Method}

\begin{figure*}[t]
\begin{center}
\includegraphics[width=\linewidth]{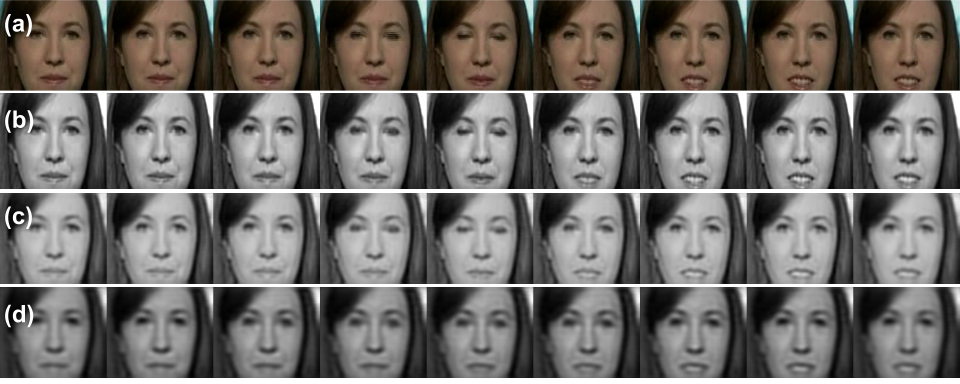}
\end{center}
   \caption{Results from reconstruction using temporal convolutional autoencoder on a face video. (a) Input video sequence. (b) Reconstruction after using 4 convolutional layers. (c) Reconstruction after using 8 layers. (d) Reconstruction after using 12 layers.}
\label{fig:autoencoder_results}
\end{figure*}

We propose a semi-supervised approach using a deep neural network, by combining an autoencoder with a classification loss function, and training both of them in parallel. The input for the first layer is a short sequence of facial gestures composed of 9 frames cropped to $145 \times 145$ pixels window. The loss function is evaluated by combining a predictive loss from 7 different pre-labeled gestures (for the labeled part of the dataset), and autoencoder Euclidean loss for the entire (labeled and un-labeled) collection. The weights of each layer are dynamically altered such that the importance of the autoencoder loss decreases with relation to the predictive loss as the training progresses. While generating the data, we use Viola and Jones face detection \cite{viola2004robust} for cropping the faces. We use slow fusion based convolutional neural network with convolutions in both space and time (see figure \ref{fig:completearch} for a detailed overview).

\subsection{Action autoencoder}\label{subsectionautoencoder}

\begin{figure*}
\begin{center}
\includegraphics[width=0.95\linewidth]{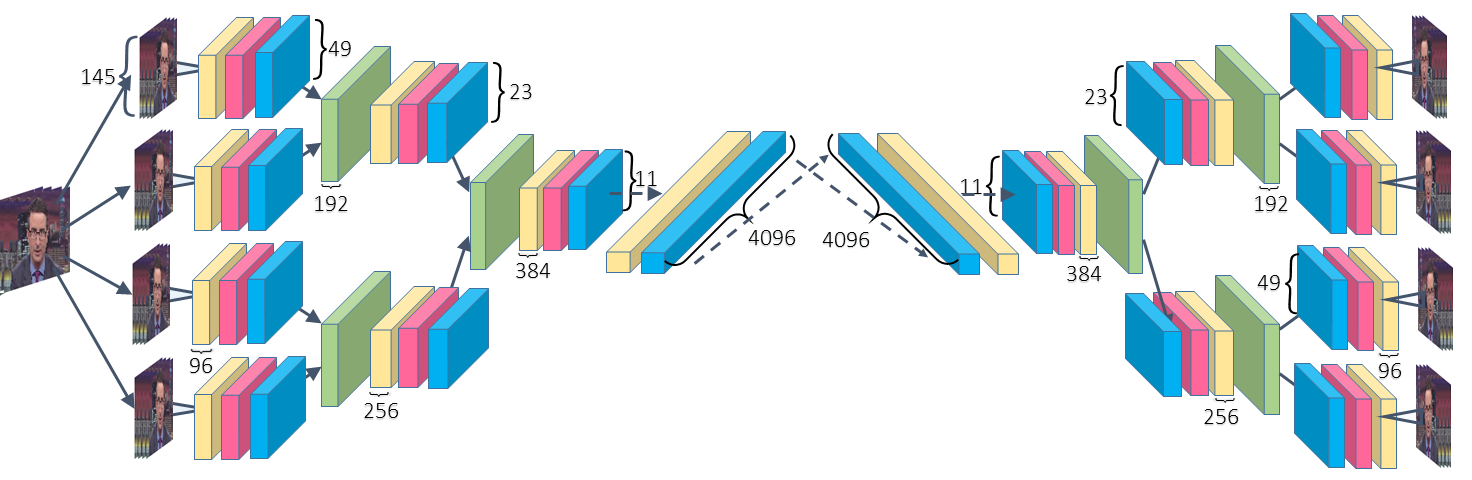}
\end{center}
   \caption{Convolution autoencoder using slow fusion technique combined with convolutions in time. Deconvolution layers \textit{deconvolve} the temporal features and reconstruct frames which are compared to the original input.}
\label{fig:autoencoder}
\end{figure*}

Our action autoencoder consists of convolutional autoencoder for learning deep features and reducing the dimensionality of the data. We use convolutional filters with weight sharing in the first 6 layers followed by 2 fully connected layers. This network is similar to Imagenet \cite{krizhevsky2012imagenet} but accepts inputs of size $145 \times 145 \times 9$ as an input. Using shorthand notation the full architecture can be written as $C(96,11,3)-N-C(256,5,2)-N-C(384,3,2)-N-FC(4096)-FC(4096)-DC(96,11,3)-N-DC(256,5,2)-N-DC(384,3,2)$, where $C(n,f,s)$ stands for convolution layers with $n$ filters of size $f$ and stride $s$. $DC(n,f,s)$ stands for deconvolution layers with $n$  deconvolving filters of size $f$ and stride $s$. $FC(n)$ stands for fully connected layers with $n$ nodes and $N$ stands for local response normalization layers. We extend the convolution layers in time and use slow fusion model \cite{karpathy2014large} which slowly combines temporal information in successive layers. The first convolution filters have size 3 and stride 2 in time domain, the next layer has size 2 and stride 2 and the third layer combines all temporal features. The deconvolution layers are extended in time as well and reverse the slow fusion generating temporal features successively (see figure \ref{fig:autoencoder}).

\subsection{Multi-Velocity Encoders}\label{subsectionmultivelocity}

One of the main challenges in action recognition is related to assigning similar classification to objects at different velocities. In this work we propose to learn the velocity of the sequence in parallel to its classification by adaptive temporal interpolation. Our multi-velocity autoencoder consists of 3 action autoencoders combined together to access temporal features for different velocities. We achieve this by adding a convolution layer as the first layer which uses cubic b-spline interpolation to \textit{slow down} the video and generate intermediate frames. Piece-wise cubic b-spline interpolation is preferred over polynomial techniques as it can minimize interpolation error for fewer points and lower degree polynomials \cite{hou1978cubic}. For initialization a sampling factor of $1$, $2/3$ and $1/3$ is chosen, which is later refined as a part of the learning.


\begin{figure*}
\begin{center}
\includegraphics[width=0.95\linewidth]{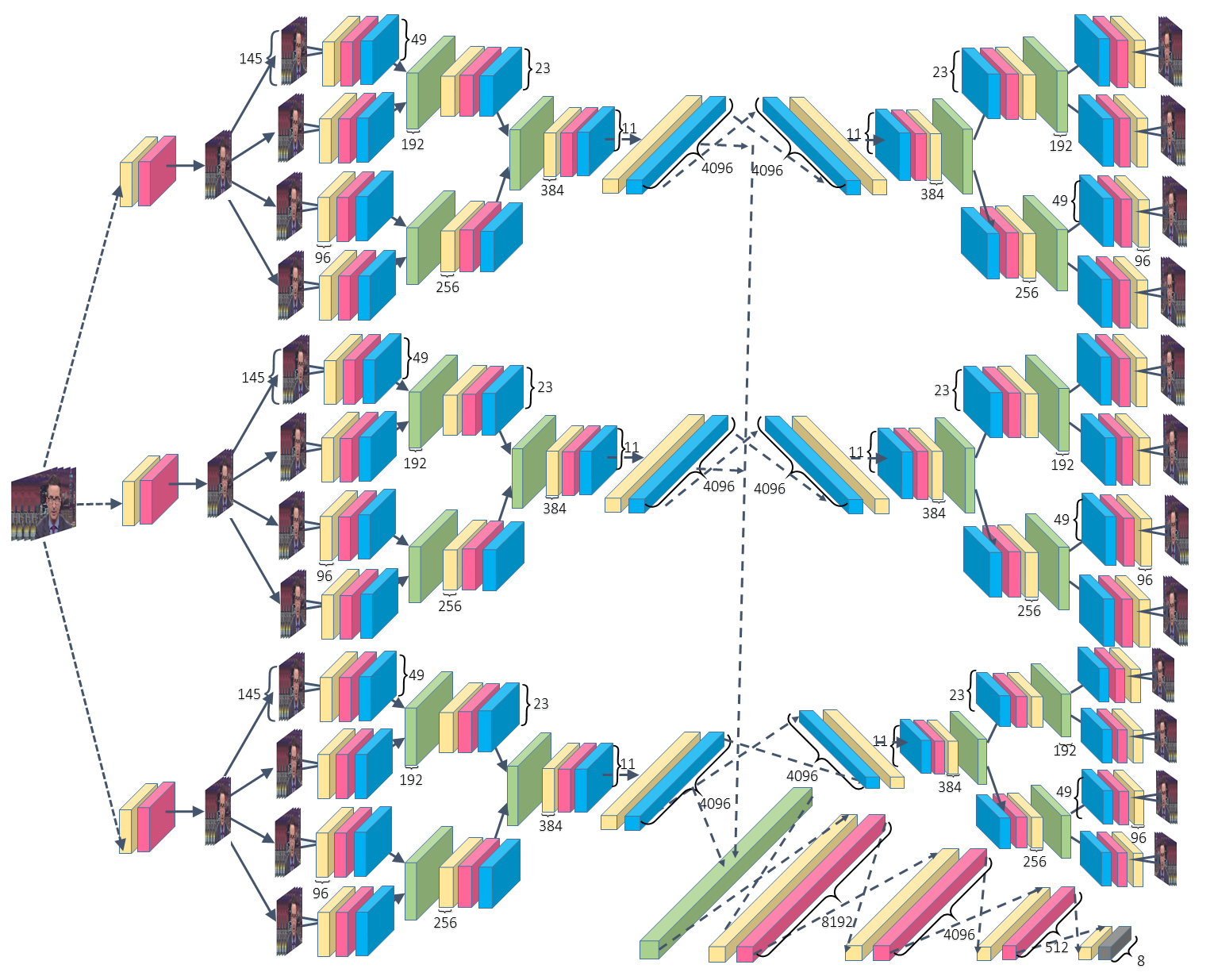}
\end{center}
   \caption{Complete architecture of multi-video semi supervised learner comprises of temporal filters for generating video frames at multiple velocities serving as input to 3 separate autoencoders. The predictor merges deep features from all 3 autoencoders and learns classification labels using deep neural net on top of these. }
\label{fig:completearch}
\end{figure*}

Next we show how to generate the required weights for interpolation and encode them as a neural network layer. Cubic b-splines are continuous \textit{piecewise-polynomial} functions containing polynomials of degree $3$ or less. A cubic b-spline spanning $n+1$ points comprises of $n$ cubic polynomials $\left( \mathbf{S_n}(x)^{N}_{n=1} \right)$ which can be uniquely defined using $4n$ coefficients. These coefficients can be recovered by applying linear constraints arising from continuity and differentiability of the function on the break points (or \textit{knots}). We represent input video at each pixel as a function of time
and use cubic b-splines to approximate intermediate values. We represent intermediate polynomials between $n+1$ frames as a coefficient vector $\bar{p}$ containing coefficients for all $n$ polynomials.

 Let $\bar{x},\bar{y}$ be the frame numbers and pixel values known to us at the different frames, these frames are obvious choices for break points as we try approximating space between frames using b-spline curves. Cubic b-spline coefficients $\bar{p}$ for each pixel can be generated by solving a linear equation $\mat{A}\bar{p} = \mat{T}\bar{y}$ as shown in the appendix. Here both $\mat{A}$ and $\mat{T}$ depend only on frame numbers ($\bar{x}$) and are independent of pixel values ($\bar{y}$) or pixel coordinates.

Let $u_o$ be one of new points in time where we want to interpolate a video frame, we can compute it now by selecting $k^{th}$ consecutive frames containing $u_o$, choosing piecewise polynomial contained in-between these frames  $\left(\mathbf{S_k}(x)\right)$ and evaluating it at $u_o$. We can write this as a dot product between coefficients and input $\bar{r} \cdot \bar{p}$, where $u_o$ lies between $k^{th}$ consecutive frames and $\bar{r}$ is as defined below in \eqref{eqn:singleinterpolant}:

\begin{equation}
\label{eqn:singleinterpolant}
\bar{[r_i]} =\begin{cases}
 u_o^{i-4k} & \text{if } x_{\floor{i/4}} \leq u_o \leq x_{\floor{i/4}+1} \\
 0 & o/w
\end{cases}
\end{equation}

Extending \eqref{eqn:singleinterpolant} to several temporal positions; Let $\bar{u} = [u_i]$ be locations in time where frames needs to be interpolated, we compute them as a matrix vector product $\mat{R}\bar{p}$. Here each row of $\mat{R} = [r_{j,i}]$ is computed using
a shifted version of the equation given above, specifically:
%
\begin{equation}
\label{eqn:interpolantmatrix}
[r_{j,i}] =\begin{cases}
 (u_j - j)^{i-4k} & \text{if } x_{\floor{i/4}} < u_j < x_{\floor{i/4}+1} \\
 0 & o/w
\end{cases}
\end{equation}

Equation \eqref{eqn:interpolantmatrix} shows how interpolated pixel values $\mathbf{F}(\bar{u})$ are linearly related to b-spline coefficients $\mat{R}\bar{p}$. Solving for $\bar{p}$ using $\bar{p} = \mat{A^{-1}}\mat{T}\bar{y}$ we infer that  $\mathbf{F}(\bar{u}) = \mat{R}\mat{A^{-1}}\mat{T}\bar{y}$. The new velocity layer weights are initialized by computing matrix $\mat{R}\mat{A^{-1}}\mat{T}$ which is independent of pixel values $\bar{y}$ and their spatial locations. We can represent this matrix as a \textit{caffe} convolution layer with shared weights, which contains $n$ filters of size $1 \times 1 \times n$ applied to all frames of video. We use algorithm \ref{bsplinealgo} to create 3 different weight matrices which interpolate sampling factors of $1$, $2/3$ and $1/3$.

\begin{algorithm}
\caption{Generate convolution layer spline weights.}\label{bsplinealgo}
\algorithmicrequire{ Frame numbers $\bar{x}$, new temporal locations $\bar{u}$} \\
\algorithmicensure{ Caffe Weight Matrix $\mat{W}$}
\begin{algorithmic}[1]
\Function{SplineWeights}{c}
\State $nSplines \gets length(\bar{x}) - 1$
\For{$(i \gets 0;i<nSplines;i++)$}
    \State $p \gets 4i$
    \State $\mat{T_{p,i}} \gets 1$
    \State $\mat{T_{p+1,i+1}} \gets 1$
    \For{$(h \gets 0;h<=1;h++)$}
    \State $s \gets p - 4h$
    \State $\mat{A_{p+h,p:p+3}} \gets [h^3,  h^2,  h,  1]$
    \State $\mat{A_{i+2,s+4:s+8}} \gets -1^{h+1}[3h^2, 2h, 1, 0] $
	\State $\mat{A_{i+3,s+4:s+8}} \gets -1^{h+1}[6h, 2, 0, 0] $
	\EndFor
\EndFor
\State $\mat{A_{i+3,i-4:i+3}} \gets [6, 0, 0, 0, -6, 0, 0, 0] $
\State $\mat{A_{i+4,0:7}} \gets [6, 0, 0, 0, -6, 0, 0, 0] $
\For{$(i \gets 0;i<length(\bar{u});i++)$}
\State $p \gets find(\bar{x},\floor{\bar{u}(i)})$
\For{$(h \gets 0;h<4;h++)$}
\State $\mat{R_{i,4p+h}} \gets (\bar{u}(i) - \bar{x}(p))^h$
\EndFor
\EndFor
\State $\mat{W} \gets \mat{RA^{-1}T}$ \\
\Return $\mat{W}$
\EndFunction
\end{algorithmic}
\end{algorithm}

\subsection{Semi-Supervised Learner}

One of the main challenges we face today for training deep neural networks is the need for large labeled datasets. The richness of data is probably one of the main reasons why neural nets report such impressive predictive results in almost every field, but it is also extremely hard to collect and label such datasets. In Semi-supervised paradigm, we assume that only a part of the data is labeled, yet we wish to utilize the knowledge hidden within the entire set. Here we combine the action autoencoder convolution layers with a softmax loss function for the labeled set.
The classifier neural net is inspired by Imagenet \cite{krizhevsky2012imagenet}, with additional fully connected layers which are shared with the autoenoder, to generate deeper classification features from the latter. The full architecture of the predictor is  $C(96,11,3)-N-C(256,5,2)-N-C(384,3,2)-N-FC(4096)-FC(8192)-FC(4096)-FC(512)-FC(8)$ with softmax layers in the end for label classification. Please refer to section \ref{subsectionautoencoder} for explanation of the architecture shorthand.

The protocol we suggest for training the net is as important as the topology itself. We begin by training the autoencoder as a sole learner from the outer layer to the inner ones. Meaning, we adaptively add layers to the autoencoder, train the neural net, and use the produced weights as initialization for the next step. This is one of the traditional approaches used to train autoencoders  \cite{hinton2006fast, carreira2005contrastive}.  Next, we use the weights for initialization of the semi-supervised net, allowing the entire net to fine tune. A key factor in training is the learning rate of the two matched learners. We begin the training using a higher learning rate for the autoencoder (with predictor layers staying fixed using zero learning rate) and end the process with increased importance to the labeled loss function. While training on the labeled data, ratio between the two varies from a factor of $10^{3}$ to a factor of $10^5$ favoring the loss layer.

\subsubsection{Multi-Velocity Semi-Supervised Learner}

\begin{figure}[t]
\begin{center}
\includegraphics[width=\linewidth]{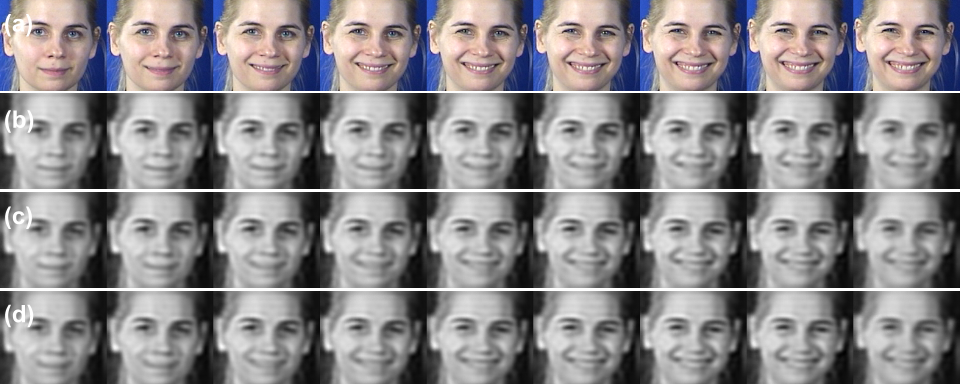}
\end{center}
   \caption{Results from reconstruction using multi velocity encoders, bottom 3 images are output from autoencoder ensemble. (a) Input video sequence. (b) Reconstruction using encoder with sampling factor of $1/3$. (c) Reconstruction using sampling factor of $2/3$. (d) Reconstruction at original velocity.}
\label{fig:multivelocity_autoencoder_results}
\end{figure}
\begin{table*}
\small
\tabcolsep=0.05cm
\begin{tabularx}{\textwidth}{l l} {
\begin{tabularx}{0.5\textwidth}{l c c c c c c c}
\multicolumn{8}{c}{}{\textbf{Confusion matrix using our methods on Cohn-Kanade}}\\
\multicolumn{1}{c}{} & \multicolumn{1}{c}{\textit{\textit{Anger}}} & \multicolumn{1}{c}{\textit{Contempt}} & \multicolumn{1}{c}{\textit{Happy}} & \multicolumn{1}{c}{\textit{Disgust}} & \multicolumn{1}{c}{\textit{Fear}} & \multicolumn{1}{c}{\textit{Sadness}}& \multicolumn{1}{c}{\textit{Surprise}}\\ 
\hhline{~~~~~~~~}
 \textit{Anger} & 0.85 \cellcolor{cmcolor!40} & 0.08 \cellcolor{blue!0} & 0 \cellcolor{blue!0} & 0 \cellcolor{blue!0} & 0 \cellcolor{blue!0} & 0.08 \cellcolor{blue!0} & 0 \cellcolor{blue!0} \\ 
\hhline{~~~~~~~~}
 \textit{Contempt} & 0 \cellcolor{blue!0} & 0.60 \cellcolor{blue!40} & 0 \cellcolor{blue!0} & 0 \cellcolor{blue!0} & 0 \cellcolor{blue!0} & 0 \cellcolor{blue!0} & 0.40 \cellcolor{blue!30} \\ 
\hhline{~~~~~~~~}
 \textit{Happy} & 0 \cellcolor{blue!0} & 0 \cellcolor{blue!0} & 1.00 \cellcolor{cmcolor!40} & 0 \cellcolor{blue!0} & 0 \cellcolor{blue!0} & 0 \cellcolor{blue!0} & 0 \cellcolor{blue!0} \\ 
\hhline{~~~~~~~~}
 \textit{Disgust} & 0.06 \cellcolor{blue!0} & 0 \cellcolor{blue!0} & 0 \cellcolor{blue!0} & 0.94 \cellcolor{cmcolor!40} & 0 \cellcolor{blue!0} & 0 \cellcolor{blue!0} & 0 \cellcolor{blue!0} \\ 
\hhline{~~~~~~~~}
 \textit{Fear} & 0 \cellcolor{blue!0} & 0 \cellcolor{blue!0} & 0.14 \cellcolor{blue!4} & 0 \cellcolor{blue!0} & 0.57 \cellcolor{blue!40} & 0 \cellcolor{blue!0} & 0.29 \cellcolor{blue!18} \\ 
\hhline{~~~~~~~~}
 \textit{Sadness} & 0.25 \cellcolor{blue!14} & 0 \cellcolor{blue!0} & 0 \cellcolor{blue!0} & 0 \cellcolor{blue!0} & 0 \cellcolor{blue!0} & 0.75 \cellcolor{cmcolor!40} & 0 \cellcolor{blue!0} \\ 
\hhline{~~~~~~~~}
 \textit{Surprise} & 0 \cellcolor{blue!0} & 0 \cellcolor{blue!0} & 0 \cellcolor{blue!0} & 0 \cellcolor{blue!0} & 0 \cellcolor{blue!0} & 0 \cellcolor{blue!0} & 1.00 \cellcolor{cmcolor!40} \\ 
\hhline{~~~~~~~~}
\multicolumn{8}{c}{}\\
\end{tabularx}
} &
{
\begin{tabularx}{0.5\textwidth}{l c c c c c c c}
\multicolumn{7}{c}{}{\textbf{Confusion matrix using external methods}}\\
\multicolumn{1}{c}{} & \multicolumn{1}{c}{\textit{Anger}} & \multicolumn{1}{c}{\textit{Contempt}} & \multicolumn{1}{c}{\textit{Happy}} & \multicolumn{1}{c}{\textit{Disgust}} & \multicolumn{1}{c}{\textit{Fear}} & \multicolumn{1}{c}{\textit{Sadness}}& \multicolumn{1}{c}{\textit{Surprise}}\\ 
\hhline{~~~~~~~~}
 \textit{Anger} & 0.73 \cellcolor{blue!40} & 0 \cellcolor{blue!0} & 0.07 \cellcolor{blue!0} & 0 \cellcolor{blue!0} & 0 \cellcolor{blue!0} & 0.20 \cellcolor{blue!9} & 0 \cellcolor{blue!0} \\ 
\hhline{~~~~~~~~}
 \textit{Contempt} & 0 \cellcolor{blue!0} & 0.86 \cellcolor{cmcolor!40} & 0 \cellcolor{blue!0} & 0 \cellcolor{blue!0} & 0 \cellcolor{blue!0} & 0.14 \cellcolor{blue!4} & 0 \cellcolor{blue!0} \\ 
\hhline{~~~~~~~~}
 \textit{Happy} & 0 \cellcolor{blue!0} & 0 \cellcolor{blue!0} & 0.95 \cellcolor{blue!40} & 0 \cellcolor{blue!0} & 0.05 \cellcolor{blue!0} & 0 \cellcolor{blue!0} & 0 \cellcolor{blue!0} \\ 
\hhline{~~~~~~~~}
 \textit{Disgust} & 0.25 \cellcolor{blue!14} & 0.12 \cellcolor{blue!2} & 0 \cellcolor{blue!0} & 0.38 \cellcolor{blue!27} & 0 \cellcolor{blue!0} & 0.12 \cellcolor{blue!2} & 0.12 \cellcolor{blue!2} \\ 
\hhline{~~~~~~~~}
 \textit{Fear} & 0 \cellcolor{blue!0} & 0 \cellcolor{blue!0} & 0 \cellcolor{blue!0} & 0 \cellcolor{blue!0} & 1.00 \cellcolor{cmcolor!40} & 0 \cellcolor{blue!0} & 0 \cellcolor{blue!0} \\ 
\hhline{~~~~~~~~}
 \textit{Sadness} & 0.33 \cellcolor{blue!23} & 0 \cellcolor{blue!0} & 0 \cellcolor{blue!0} & 0.11 \cellcolor{blue!1} & 0 \cellcolor{blue!0} & 0.44 \cellcolor{blue!34} & 0.11 \cellcolor{blue!1} \\ 
\hhline{~~~~~~~~}
 \textit{Surprise} & 0 \cellcolor{blue!0} & 0 \cellcolor{blue!0} & 0 \cellcolor{blue!0} & 0.05 \cellcolor{blue!0} & 0 \cellcolor{blue!0} & 0 \cellcolor{blue!0} & 0.95 \cellcolor{blue!40} \\ 
\hhline{~~~~~~~~}
\multicolumn{8}{c}{}\\
\end{tabularx}
} \\
\end{tabularx}
\caption{Confusion matrices over test results for Cohn Kanade dataset using our methods and best performing external method which uses \textit{Expressionlets} \cite{liu2014learning}.  On the left we show results for the proposed multi-velocity semi-supervised approach across various facial expressions, while on the right we present confusion matrix from Expressionlets method. Highest accuracy in each category is marked using green color. We outperform competing method in 5 verticals by getting 100\% accuracy on happiness, 100\% on surprise, 94\% on disgust, 85\% in anger and 75\% in sadness. For both methods misclassification occur when emotions like sadness get recognized as anger and \textit{vice-versa}.}
\label{table:cfckplus}
\end{table*}

Finally we attach the new proposed Multi-Velocity layers as the first structure of the semi-supervised neural net. Each sub-structure (See Figure \ref{fig:completearch}), has its own autoencoder, all of which are concatenated after the inner most convolution layer into a feature vector (size $12288$), later used by the labeled loss function.  The learner loss function can be expressed as a weighted sum of autoencoder and predictor loss given in equation \ref{predictorloss} below.

\begin{equation}
\label{predictorloss}
L = \alpha\sum_v{||\bar{x}-\bar{x}_v||} - \beta\sum_j{y_j log \left(\frac{e^{o_j}}{\sum_k{e^{o_k}}}\right)}
\end{equation}

Here $\bar{x}, \bar{x}_v$ are autoencoder inputs and outputs, $y_j$ are the input labels and $o_j$ is the outputs from predictor layer. $\sum_v{||\bar{x}-\bar{x}_v||}$ is the combined Euclidean loss across three multi-velocity encoders and $- \sum_j{y_j  log \left(\frac{e^{o_j}}{\sum_k{e^{o_k}}}\right)}$ is softmax loss \cite{bengio2005convex}. While training using labeled data, the loss coefficient $\beta$ is selected to keep softmax loss an order of magnitude higher than the Euclidean loss. Loss coefficient $\alpha$ is adjusted as softmax loss goes down to continue training predictor layers, without overfitting autoencoder layers. Notice that we use two coefficients for the energy function and not just controlling the ratio between the two since the back-propagation algorithm has its own additional parameters.

\section{Datasets}
In order to evaluate the proposed architecture we use two known datasets from literature as well as present two additional datasets collected by us; The first dataset contains more than 160 million images combined into 6.5 million short (25 frames) clips, used by us to train our autoencoders. The second dataset is comprised of 2777 short clips labeled for seven emotions.
In the following section we elaborate on the four datasets.

\subsection{Autoencoder dataset}
In order to train very deep neural nets we must obtain a huge collection of data. Here we collected 6.5 million video clips containing 25 frames each, adding up to more than 162 million face images. We used viola-jones face detector to find and segment out the faces. Next, we localized landmarks for each frame using a deformable model for the face \cite{asthana2014incremental} and detected the facial pose by fitting a 3D model to the landmarks. This process allowed us to restrict the dataset to videos which contain faces tilted less than 30 degrees and remove any faces looking sideways.

In order to extract only meaningful video clips we removed clips with static gestures or those where the faces were rapidly altering, either due to some high speed movement or simply due to appearance of a different face. We achieved this by blurring the clips and calculating the difference between consecutive frames.

The raw videos were taken from public sources such as CNN, MSNBC, FOX and CSPAN. To our knowledge this is the biggest facial dataset reported in literature, and we plan to make it public.

\subsection{Asevo dataset}
In order to collect and label our own gestures  we developed a video recording and annotation tools. We developed the application using python based OpenCV and captured the clips using Logitech C920 HD camera. The database contains facial clips from 160 subjects both male and female, where gestures were artificially generated according to a specific request, or genuinely given due to a shown stimulus. We collected a total of 2777 clips out of which 1745 were captured after providing the stimulus while 1032 were generated artificially.  To create natural facial expressions we selected a bank of YouTube videos for each facial expression and showed them to subjects, capturing their reaction to the visual stimulus.
We quantitatively summarize this dataset in table \ref{table:Asevodistribution}, where posed clips refers to the artificially generated expressions and non-posed to the stimulus activation procedure.

\begin{table}
\begin{center}
\begin{tabularx}{0.95\linewidth}{|l|>{\centering\arraybackslash}X|>{\centering\arraybackslash}X|>{\centering\arraybackslash}X|}

\hline
Emotion & Posed & Non-Posed & Cumulative \\
\hline\hline
Anger & 132 & 318 & 450 \\
Sadness & 118 & 148 & 266 \\
Contempt & 153 & 301 & 454 \\
Fear & 137 & 96 & 233 \\
Surprise & 188 & 232 & 420 \\
Joy & 172 & 503 & 675 \\
Disgust & 132 & 147 & 279 \\
\hline
Total & 1032 & 1745 & 2777 \\
\hline
\end{tabularx}
\end{center}
\caption{Data distribution for Asevo dataset for various emotions. Posed clips refer to the artificially generated clips, while non-posed refer to those captured using the stimulus activation procedure.}
\label{table:Asevodistribution}
\end{table}

\subsection{Cohn Kanade Dataset}

\begin{table*}
\small
\tabcolsep=0.05cm
\begin{tabularx}{\textwidth}{l l} {
\begin{tabularx}{0.5\textwidth}{l c c c c c c c}
\multicolumn{8}{c}{}{\textbf{Confusion matrix using our methods on Asevo Dataset}}\\
\multicolumn{1}{c}{} & \multicolumn{1}{c}{\textit{Anger}} & \multicolumn{1}{c}{\textit{Contempt}} & \multicolumn{1}{c}{\textit{Happy}} & \multicolumn{1}{c}{\textit{Disgust}} & \multicolumn{1}{c}{\textit{Fear}} & \multicolumn{1}{c}{\textit{Sadness}}& \multicolumn{1}{c}{\textit{Surprise}}\\ 
\hhline{~~~~~~~~}
 \textit{Anger} & 0.62 \cellcolor{cmcolor!40} & 0.13 \cellcolor{blue!3} & 0.02 \cellcolor{blue!0} & 0.10 \cellcolor{blue!0} & 0.01 \cellcolor{blue!0} & 0.06 \cellcolor{blue!0} & 0.06 \cellcolor{blue!0} \\ 
\hhline{~~~~~~~~}
 \textit{Contempt} & 0.16 \cellcolor{blue!5} & 0.33 \cellcolor{blue!22} & 0.22 \cellcolor{blue!12} & 0.04 \cellcolor{blue!0} & 0.04 \cellcolor{blue!0} & 0.05 \cellcolor{blue!0} & 0.16 \cellcolor{blue!5} \\ 
\hhline{~~~~~~~~}
 \textit{Happy} & 0.01 \cellcolor{blue!0} & 0.06 \cellcolor{blue!0} & 0.86 \cellcolor{cmcolor!40} & 0.01 \cellcolor{blue!0} & 0.00 \cellcolor{blue!0} & 0.03 \cellcolor{blue!0} & 0.03 \cellcolor{blue!0} \\ 
\hhline{~~~~~~~~}
 \textit{Disgust} & 0.25 \cellcolor{blue!14} & 0.13 \cellcolor{blue!2} & 0.16 \cellcolor{blue!6} & 0.31 \cellcolor{cmcolor!30} & 0.02 \cellcolor{blue!0} & 0.06 \cellcolor{blue!0} & 0.07 \cellcolor{blue!0} \\ 
\hhline{~~~~~~~~}
 \textit{Fear} & 0.15 \cellcolor{blue!5} & 0.03 \cellcolor{blue!0} & 0.03 \cellcolor{blue!0} & 0.10 \cellcolor{blue!0} & 0.18 \cellcolor{cmcolor!22} & 0.04 \cellcolor{blue!0} & 0.46 \cellcolor{blue!36} \\ 
\hhline{~~~~~~~~}
 \textit{Sadness} & 0.23 \cellcolor{blue!13} & 0.10 \cellcolor{blue!0} & 0.05 \cellcolor{blue!0} & 0.04 \cellcolor{blue!0} & 0.06 \cellcolor{blue!0} & 0.18 \cellcolor{cmcolor!22} & 0.34 \cellcolor{blue!24} \\ 
\hhline{~~~~~~~~}
 \textit{Surprise} & 0.06 \cellcolor{blue!0} & 0.05 \cellcolor{blue!0} & 0.02 \cellcolor{blue!0} & 0.04 \cellcolor{blue!0} & 0.08 \cellcolor{blue!0} & 0.09 \cellcolor{blue!0} & 0.66 \cellcolor{cmcolor!40} \\ 
\hhline{~~~~~~~~}
\multicolumn{8}{c}{}\\
\end{tabularx} 
} &
{
\begin{tabularx}{0.5\textwidth}{l c c c c c c c}
\multicolumn{7}{c}{}{\textbf{Confusion matrix using external methods}}\\
\multicolumn{1}{c}{} & \multicolumn{1}{c}{\textit{Anger}} & \multicolumn{1}{c}{\textit{Contempt}} & \multicolumn{1}{c}{\textit{Happy}} & \multicolumn{1}{c}{\textit{Disgust}} & \multicolumn{1}{c}{\textit{Fear}} & \multicolumn{1}{c}{\textit{Sadness}}& \multicolumn{1}{c}{\textit{Surprise}}\\ 
\hhline{~~~~~~~~}
 \textit{Anger} & 0.61 \cellcolor{blue!40} & 0.08 \cellcolor{blue!0} & 0.06 \cellcolor{blue!0} & 0.07 \cellcolor{blue!0} & 0.04 \cellcolor{blue!0} & 0.06 \cellcolor{blue!0} & 0.08 \cellcolor{blue!0} \\ 
\hhline{~~~~~~~~}
 \textit{Contempt} & 0.09 \cellcolor{blue!0} & 0.44 \cellcolor{cmcolor!34} & 0.27 \cellcolor{blue!17} & 0.06 \cellcolor{blue!0} & 0.02 \cellcolor{blue!0} & 0.06 \cellcolor{blue!0} & 0.05 \cellcolor{blue!0} \\ 
\hhline{~~~~~~~~}
 \textit{Happy} & 0.01 \cellcolor{blue!0} & 0.07 \cellcolor{blue!0} & 0.85 \cellcolor{blue!40} & 0.01 \cellcolor{blue!0} & 0 \cellcolor{blue!0} & 0.01 \cellcolor{blue!0} & 0.06 \cellcolor{blue!0} \\ 
\hhline{~~~~~~~~}
 \textit{Disgust} & 0.19 \cellcolor{blue!8} & 0.14 \cellcolor{blue!4} & 0.25 \cellcolor{blue!14} & 0.22 \cellcolor{blue!12} & 0.01 \cellcolor{blue!0} & 0.02 \cellcolor{blue!0} & 0.16 \cellcolor{blue!6} \\ 
\hhline{~~~~~~~~}
 \textit{Fear} & 0.21 \cellcolor{blue!10} & 0.09 \cellcolor{blue!0} & 0.06 \cellcolor{blue!0} & 0.07 \cellcolor{blue!0} & 0.12 \cellcolor{blue!2} & 0.06 \cellcolor{blue!0} & 0.39 \cellcolor{blue!29} \\ 
\hhline{~~~~~~~~}
 \textit{Sadness} & 0.26 \cellcolor{blue!15} & 0.20 \cellcolor{blue!9} & 0.19 \cellcolor{blue!8} & 0.02 \cellcolor{blue!0} & 0.05 \cellcolor{blue!0} & 0.13 \cellcolor{blue!2} & 0.16 \cellcolor{blue!6} \\ 
\hhline{~~~~~~~~}
 \textit{Surprise} & 0.15 \cellcolor{blue!4} & 0.03 \cellcolor{blue!0} & 0.09 \cellcolor{blue!0} & 0.01 \cellcolor{blue!0} & 0.06 \cellcolor{blue!0} & 0.05 \cellcolor{blue!0} & 0.61 \cellcolor{blue!40} \\ 
\hhline{~~~~~~~~}
\multicolumn{8}{c}{}\\
\end{tabularx}
}
\end{tabularx}
\caption{Confusion matrix over test results for Asevo dataset using the proposed multi-velocity semi-supervised learner (left) and external competing method using \textit{Covariance Riemann kernel} \cite{liu2014combining}. We outperform multiple kernel based methods in 6 out of 7 emotion categories. Similar to the Cohn-Kanade dataset (see table \ref{table:cfckplus}), both methods fail when disgust, fear or sadness get classified as anger.}
\label{table:cfAsevo}
\end{table*}

The Cohn Kanade Dataset \cite{lucey2010extended} is one of the most popular datasets used for facial expression recognition. The dataset contains 593 sequences out of which 327 are labeled for 7 emotions. Along with posed facial expressions, the dataset also contains non-posed smile expressions. However the dataset lacks depth in having other non-posed expressions and is not extensive as Asevo dataset in capturing naturally expressed emotions. Each video clip contains facial expression going from baseline neutral to peak of expressed emotion.


\begin{table*}
\begin{center}
\begin{tabular}{|>{\hspace{1pt}}l<{\hspace{45pt}}|>{\hspace{7pt}}c<{\hspace{7pt}}|>{\hspace{7pt}}c<{\hspace{7pt}}|>{\hspace{7pt}}c<{\hspace{7pt}}|}
\hline
\textbf{Dataset} & \textbf{\textit{MMI}} & \textbf{\textit{CKPLUS}} & \textbf{\textit{Asevo}} \\
\hline
\textbf{Technique} & \multicolumn{3}{|c|}{} \\
\hline
External Methods & \multicolumn{3}{|c|}{} \\
\hline
\textit{Covariance riemann kernel based multiple kernel methods} & 40.9 & 79  & \cellcolor{yellow!20}51.05  \\
\hline
\textit{Multiple kernel methods with gaussian riemann kernel} & 40.9 & 67 & 46.92 \\
\hline
\textit{Grassman kernel based multiple kernel approaches} & 9.09 & 17.9 & 44.99 \\
\hline
\textit{Expressionlets based manifold learning techniques} & \cellcolor{yellow!20}52.91 & \cellcolor{yellow!20}82.7 & 48.6 \\
\hline
\multicolumn{4}{|c|}{} \\
\hline
Our Methods & \multicolumn{3}{|c|}{} \\
\hline
\textit{Semi-Supervised Learner for gesture classification} & \cellcolor{cmcolor!42}59.01 & 87.36 & 51.11 \\
\hline
\textit{Multi-Velocity Encoder based learner for gesture classification} & 58.7 & \cellcolor{cmcolor!42}89.47 & \cellcolor{cmcolor!42}52.59 \\
\hline
\end{tabular}
\end{center}
\caption{Comparison of results from various techniques on CKPLUS, MMI and Asevo datasets. The dataset was divided into 3 parts test, train and val randomly. Training set was 50\%, test and validation were 30\% and 20\% respectively. Our method consistently won for both small and large datasets (winning method is shown in green and the leading method is showed using yellow).}
\label{table:predictor_results}
\end{table*}

\subsection{MMI Dataset}

MMI facial expression dataset \cite{pantic2005web} is an ongoing effort for representing both posed and non posed facial expressions. The dataset has total 2894 video clips out of which 197 have been labeled for six basic emotions. MMI originally contained only posed facial expressions and recently was extended to contain induced happiness, disgust and surprise \cite{valstar2010induced}. Each video clip in MMI contains people going from neutral to peak and then back to neutral facial expression.

\section{Experiments and Results}\label{experimentsandresults}

\subsection{Video autoencoder}
Our first experiment shows qualitatively results  of a single video autoencoder. We use $145 \times 145 \times 9$ clips as input, where the spatial resolution received by downsampling all clips to that single size using bspline interpolation, and  $9$ frames are extracted from the clip by using every third frame.  We use \textit{caffe} \cite{jia2014caffe} to train the system. In practice we convert each video clip into a image strip containing consecutive frames placed horizontally and use  \textit{caffe "imagedata", "split"} and \textit{"concat"} layers for video data input.

We minimize contrastive divergence \cite{carreira2005contrastive} to train autoencoder layers \textit{successively}. We train the first 4 beginning and end layers by creating an intermediate neural network $(C(96,11,3)-N-C(256,5,2)-N-DC(256,5,2)-N-DC(384,3,2))$ and training it on facial video clips. We then train third convolution and deconvolution layer by initializing weights from previously trained neural net and fixing the weights for first 4 beginning and end layers. We fine tune all layers once the neural net weights have converged. We repeat the process for fourth fully connected layer to generate deep features.

Please refer to figure \ref{fig:autoencoder_results} to see results from neural net based reconstruction using different number of layers.

\subsection{Multi-velocity video autoencoder}
Multi velocity semi-supervised learner comprises of an array of three independent autoencoders and a predictor net. We initialize the autoencoders using the weights from the video autoencoder and add a convolution layer as described in section \ref{subsectionmultivelocity}. We fine tune the multi-velocity layers by creating 3 datasets containing video clips at different velocities. We achieve that by selecting every third frame to create set 1 \textit{(speed = 3x)}, selecting every second frame to generate set 2 \textit{(speed = 2x)} and taking first 9 frames for set 3 \textit{(speed = 1x)}.
The weights from this step are used for initialization of our multi-velocity predictor which described next.

\subsection{Multi-velocity predictor}

For training, testing and validation we divide each dataset into 3 parts randomly. We select 50\% inputs for training, 30\% of dataset for testing and use 20\% of dataset for validation.
After the dataset was split, we further increased the size of the training dataset by shifting each video along both axes, rotating images and taking their mirror.

We train our proposed semi-supervised learner and the multi-velocity semi-supervised learner on the three datasets (MMI, CK and Asevo), and
 compare our results against multiple kernel methods \cite{liu2014combining} and expression-lets base approaches \cite{liu2014learning}. We used sources downloaded from \textit{Visual Information Processing and Learning Resources} \cite{vipl} as a reference to compare to our methods.  Note that we made the same data partitioning scheme (train, validation, test) for all methods to show a fair comparison.

We outperform all the methods compared on all the datasets used, by a substantial gap, in almost all cases.
We summarize our findings in Table \ref{table:predictor_results}, and show confusion matrices per facial expression in Tables  \ref{table:cfckplus} and \ref{table:cfAsevo}. \red{For baseline comparison against other deep neural architectures, we compare our methods against \cite{krizhevsky2012imagenet} and GoogleNet \cite{szegedy2015going}. We further verified our results against prior state of the art methods discussed in \cite{liu2014deeply} by performing \textbf{10 fold cross validation}. On MMI we get 66.15 (vs 63.4) \% and on CK+ we get 94.18 (vs 92.4) \%, making our method \textbf{state of the art} for face expression recognition.}

\section{Discussion and Future Work}

This paper presents a learning strategy for large datasets with a dramatically lower number of labeled points, in addition to new layers carefully designed to improve recognition in multi-velocity setup. We currently trim the videos to the facial window using Viola and Jones face detection, and focus solely on frontal views. Recognition in-the-wild still remains a challenge with a known low success rate. We believe that given a large and rich dataset this problem would be feasible to solve in our system, and we plan to explore that in the future.

We introduced a new layer, which adaptively resamples the videos, achieving a multi-velocity invariant learning procedure. Inserting invariants into a learning process is a research direction that we must push forward. Today training of deep neural network is still time consuming, where huge clusters are being heavily used on reasonably large datasets. We are already reaching the time-space limit of this process, and better/smarter approaches need to be considered for advancement. Our multi-velocity setup is one approach for reducing the need for data in multiple velocities, while other invariants should be explored in future work.

\section{Conclusions}

In this paper we introduced a new topology and learning protocol for semi-supervised convolutional neural networks on video sequences. We further developed a multi-velocity layer based on temporal resampling which was tuned as part of the learning procedure on an enormous collected facial dataset. We report state-of-the-art results on our own data and on public available datasets.

\section*{Appendix A: Equations for B spline Interpolation}\label{appendixbspline}

Let $\{x_k,f(x_k)\}^N_{k=0}$ be $N+1$ observations of a function $f$. Cubic spline is defined as a set of polynomials $\mathbf{S_n}(x)^{N-1}_{n=0}$ with coefficients $p_{n,i}$ which approximate $f$ as follows

\begin{equation}
\label{splineexp}
\begin{aligned}
\mathbf{S}(x) \hspace{0.35em}= \hspace{0.35em} \mathbf{S_n}(x) \hspace{0.35em}= \hspace{0.35em} p_{n,0} \hspace{0.35em}+ \hspace{0.35em} p_{n,1}(x_n-x_k) \hspace{0.35em}+ \hspace{0.35em} \\
 p_{n,2}(x_n-x_k)^2 \hspace{0.35em}+ \hspace{0.35em} p_{n,3}(x_n-x_k)^3,
\end{aligned}
\end{equation}
where $x_{k}<x_n<x_{k+1}$. We need at least $4N$ constraints to recover $p_{n,i}$ uniquely. We can generate $4N - 2$ constraints by fixing the values of polynomials at the boundaries and assuming first and second derivatives of adjacent polynomials coincide at the boundaries as well. We  add additional constraints assuming that the curve is \textit{natural} \cite{chung1979spline} and has zero derivative at boundaries. The coefficients $p_{k,n}$ are constrained by:

\begin{subequations}
\label{splinecons}
\begin{flalign}
& \mathbf{S}(x) = f(x_k) &\forall k \in \{0..N\}\\
& \mathbf{S_k}(x) - \mathbf{S_{k+1}}(x_{k+1}) = 0 &\forall k \in \{0..N-2\}\\
& \mathbf{S^{\prime}_k}(x) - \mathbf{S^{\prime}_{k+1}}(x_{k+1}) = 0 &\forall k \in \{0..N-2\}\\
& \mathbf{S^{\prime\prime}_k}(x) - \mathbf{S^{\prime\prime}_{k+1}}(x_{k+1})= 0&\forall k \in \{0..N-2\} \\
& \mathbf{S^{\prime\prime}_0}(x) = \mathbf{S^{\prime\prime}_{N-1}}(x) = 0
\end{flalign}
\end{subequations}

Let $\mat{A}$ denote matrix representing the constraints on spline polynomial coefficients and $\bar{p}$ 
 represent coefficients as described in equation \ref{splinecons}. Let $\bar{y}$ be the vector of function values $f(x_k), k \in \{0..N\} $ known to us. Right side of \ref{splinecons} can be written as a product of matrix $\mat{T}$ with vector $\bar{y}$, where $\mat{T}$ is a binary matrix. Then $\mat{A}\bar{p} = \mat{T}\bar{y}$ or $\bar{p} = \mat{A}^{-1}\mat{T}\bar{y}$.

{\small
\bibliographystyle{ieee}
\bibliography{video_expression_paper}
}

\end{document}